\documentclass{article}
\pdfoutput=1
\usepackage{ijcai22}

\usepackage{times}
\usepackage{soul}
\usepackage{url}
\usepackage[hidelinks]{hyperref}
\usepackage[utf8]{inputenc}
\usepackage[small]{caption}
\usepackage{graphicx}
\usepackage{amsmath}
\usepackage{amsthm}
\usepackage{booktabs}
\usepackage{algorithm}
\usepackage{algorithmic}
\usepackage{multirow}
\usepackage{booktabs}
\usepackage{bbding}
\usepackage{makecell}
\usepackage{xspace}
\usepackage{amssymb}

\usepackage{dcolumn}
\newcolumntype{d}[1]{D{.}{.}{#1}}% or D{.}{,}{#1} or D{.}{\cdot}{#1}

\newcommand{\eat}[1]{}
\newcommand{\paratitle}[1]{\vspace{1ex}\noindent \textbf{#1}}

\renewcommand{\vec}[1]{\mathbf{#1}}

\newcommand{\eg}{\emph{e.g.,}\xspace}

\newcommand{\ie}{\emph{i.e.,}\xspace}
\newcommand{\etc}{\emph{etc.}\xspace}

\title{Declaration-based Prompt Tuning for Visual Question Answering}

\author{
Yuhang Liu$^{1,2}$
\and
Wei Wei$^{1,2}$\footnote{
    Corresponding author.
}\and
Daowan Peng$^{1,2}$\And
Feida Zhu$^3$
\affiliations
$^1$Cognitive Computing and Intelligent Information Processing (CCIIP) Laboratory, School of Computer Science and Technology, Huazhong University of Science and Technology, China \\
$^2$Joint Laboratory of HUST and Pingan Property \& Casualty Research (HPL), China\\
$^3$School of Computing and Information Systems, Singapore Management University, Singapore
\emails
\{lyuhang, weiw, pengdw\}@hust.edu.cn,
fdzhu@smu.edu.sg
}
\begin{document}

\maketitle

\begin{abstract}
    In recent years, the \emph{pre-training-then-fine-tuning paradigm} has yielded immense success on a wide spectrum of cross-modal tasks, such as visual question answering (VQA), in which a visual-language (VL) model is first optimized via self-supervised task objectives, \eg masked language modeling (MLM) and image-text matching (ITM), and then fine-tuned to adapt to downstream task (\eg VQA) via a brand-new objective function, \eg answer prediction. However, the inconsistency of the objective forms not only severely limits the generalization of pre-trained VL models to downstream tasks, but also requires a large amount of labeled data for fine-tuning. To alleviate the problem, we propose an innovative VL fine-tuning paradigm (named \underline{\textbf{D}}eclaration-based \underline{\textbf{P}}rompt \underline{\textbf{T}}uning, abbreviated as DPT), which fine-tunes the model for downstream VQA using the pre-training objectives, boosting the effective adaptation of pre-trained models to the downstream task. Specifically, DPT reformulates the VQA task via (1) \textit{textual adaptation}, which converts the given questions into declarative sentence form for prompt-tuning, and (2) \textit{task adaptation}, which optimizes the objective function of VQA problem in the manner of pre-training phase. Experimental results on GQA dataset show that DPT outperforms the fine-tuned counterpart by a large margin regarding accuracy in both fully-supervised (2.68\%) and zero-shot/few-shot (over 31\%) settings. The data and codes are available at \url{https://github.com/CCIIPLab/DPT}.
\end{abstract}

\section{Introduction}

% (1) 从大规模VL预训练模型引出
% (2) 阐述这种预训练模型的策略

Recently, large-scale vision-language pre-training has been an emerging topic in the multi-modal community, and delivered strong performance in numerous vision-language tasks \cite{Yao2021CPTCP,Li2020OscarOA,Zhang2021VinVLRV,Chen2020UNITERUI,Lu2019ViLBERTPT,Su2020VLBERTPO}. Typically, a commonly-used practice is to follow the \emph{pre-training-then-fine-tuning paradigm} \cite{Liu2021PretrainPA}, in which a generic Transformer \cite{vaswani2017attention} is pre-trained on large-scale image-text datasets in a self-supervised manner, and then adapted to different downstream tasks by introducing additional parameters and fine-tuning using task-specific objectives, \eg auxiliary fully-connected layer for answer classification in visual question answering. This paradigm has greatly pushed forward the state-of-the-art of VQA task.

% 引入Example
\begin{figure}[t]
    \centering
    \includegraphics[width=1.\columnwidth]{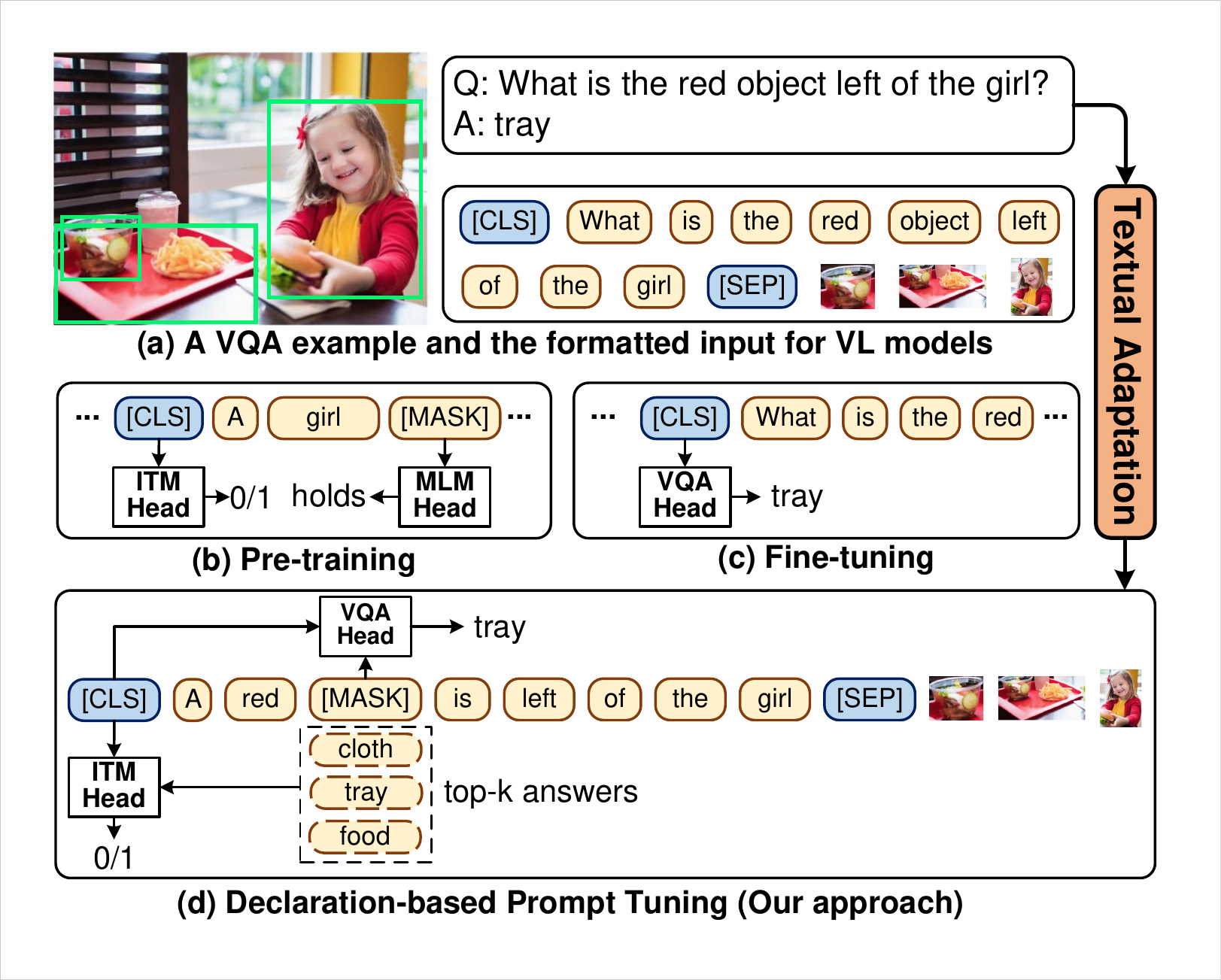} % Reduce the figure size so that it is slightly narrower than the column. Don't use precise values for figure width.This setup will avoid overfull boxes.
    \caption{ Illustration of (a) a VQA example and the formatted input for VL models, (b) pre-training VL models with masked language model (MLM) and image-text matching (ITM) tasks, (c) vanilla fine-tuning for VQA with a new classification head, and (d) our proposed declaration-based prompt tuning (DPT) framework that reformulates VQA task into fill-in-the-blank and image-text matching problems via textual and task adaptation. Only parts of the relevant image regions are shown for illustration.
    }
\label{fig:sample}
\end{figure}

% 开始引出这种模式的缺点
% 使用样例说明（预训练 vs 微调）两阶段差异
% 最后总结这种方式存在gap，导致效果不佳（主要原因就是上下游所需的能力不一致）
% 上下游任务的不同导致模型的能力存在差异
Despite the promising performance achieved, it's worth noting that there exists a natural gap in objective forms between pre-training and fine-tuning stages. As illustrated by Figure \ref{fig:sample}(b-c), most VL models are pre-trained via masked language modeling and image-text matching objectives, \ie recovering the masked token on the cross-modal contexts and predicting the matching scores of image-text pairs. However, in the fine-tuning stage, VQA problem is usually conducted and optimized using a brand-new task objective, \ie classifying [CLS] token into the semantic labels (\ie answers), where additional parameters are typically introduced. As a result, there exist great disparities in the task forms between pre-training and fine-tuning. This gap hinders the generalization of pre-trained VL models to downstream VQA task, which leads to suboptimal performance and a demand for large amount of labeled data for fine-tuning.

% 介绍这篇工作的motivation、idea
Inspired by the recent progress of vision-language pre-trained models (VL-PTM) \cite{Li2020OscarOA,Zhang2021VinVLRV} and prompt tuning paradigms in cross-modal domain \cite{Yao2021CPTCP,Tsimpoukelli2021MultimodalFL,Radford2021LearningTV}, in this paper we propose \underline{\textbf{D}}eclaration-based \underline{\textbf{P}}rompt \underline{\textbf{T}}uning (DPT), a novel paradigm of fine-tuning VL-PTM for VQA problem. Our core insight is to reformulate the objective form of downstream VQA task into the format of pre-training phase, maximally mitigating the gap between two stages. To achieve this goal, we reformulate the VQA task from two aspects (refer to Figure \ref{fig:sample}(d)): (1) \textit{textual adaptation} that converts the textual input (\ie questions) into declarative sentence form, and (2) \textit{task adaptation} that solves VQA by recovering the masked token from the declarative sentences, and selecting the one that best matches the image. In this way, answer prediction can be achieved via cloze-filling and image-text matching, imitating the behavior of MLM and ITM tasks in the pre-training phase.

% 介绍这个Idea带来的效果
% TODO
% 在实验部分全部完成后需要更新和完善该部分
% (1) 从GQA数据集角度来说，采用Prompt-tuning实现了更好的效果+2.8%
% (2) 从泛化性来说，采用Prompt-tuning实现了不同数据集、不同VL模型的稳定和一致提升
% (3) 从数据量角度来说，在少量样本的情况下相比于原始微调方法提升更加明显（待验证）
By mitigating the gap between pre-training and fine-tuning, DPT enables strong performance over various VL models and VQA datasets in both fully-supervised and zero/few-shot settings. For example, with respect to the accuracy, our method achieves 2.68\% absolute improvement in the fully-supervised setting, and 31.8\%$\sim$37.4\% absolute improvement in the zero-shot/few-shot settings in GQA evaluation. Furthermore, the generalization experiment on VQA v2.0 equipped with recently proposed VL models shows 0.45\%$\sim$1.01\% absolute improvement compared to the vanilla fine-tuning approach.

% 总结contributions
In summary, the main contributions are the following,
\begin{itemize}
    \item We introduce \underline{\textbf{D}}eclaration-based \underline{\textbf{P}}rompt \underline{\textbf{T}}uning (DPT), a novel fine-tuning paradigm that solves VQA via adapting downstream problem to pre-training task format. To the best of our knowledge, this is the first attempt in the prompt tuning using declaration sentences for visual question answering.
    \item We propose novel textual and task adaptation approaches to reformulate VQA into cloze-filling and image-text matching problems, \ie MLM and ITM. The adapted tasks significantly outperform the fine-tuning counterparts in fully-supervised and few-shot settings.
    \item We conduct comprehensive experiments over various VL-PTMs and VQA datasets, which demonstrates the effectiveness and generalizability of DPT.
\end{itemize} 

\section{Related Work}

% 主要从两个方面来阐述相关工作：
% 多模态预训练模型 -> 预训练任务 -> 下游任务的迁移学习
% Prompt

\subsection{Pre-trained Vision-language Models}
Recently, there exists numerous work on training generic models for various downstream cross-modal tasks \cite{liu2021spatiotemporal}, such as visual question answering (VQA) or image caption \cite{Cho2021UnifyingVT,Radford2021LearningTV,Kim2021ViLTVT,Zhang2021VinVLRV,Li2020OscarOA,cheng2020stack,Tan2019LXMERTLC}. Typically, a commonly-used practice is to follow a paradigm from model pre-training to model fine-tuning. In specific, in pre-training stage, a BERT-like architecture \cite{Devlin2019BERTPO} is first built for pre-training in learning multi-modal representations via a variety of self-supervised tasks, for instance, a mask language model (MLM) task of recovering the masked textual tokens in the multi-modal context \cite{Tan2019LXMERTLC,Li2020OscarOA}, \emph{or} an image-text matching (ITM) task to verify the alignment of an image to a given text \cite{Tan2019LXMERTLC,Zhang2021VinVLRV}. Next, in the fine-tuning stage, the pre-trained model is then fine-tuned to adapt to downstream tasks using totally different task-specific objectives, such as predicting the answer for the VQA task. In this work, instead of optimizing brand-new task objectives in the \textit{fine-tune} stage, we attempt to reformulate VQA into the pre-training format, boosting the effective generalization of pre-trained VL models to the downstream task.

% Prompt的起源
% 应用于跨模态任务上
% 提到VQA任务上
\subsection{Cross-modal Prompt Tuning}
Recently, prompt tuning has increasingly received attentions due to its powerful capability in keeping the optimization objectives of the pre-trained model and the downstream task consistent \cite{Liu2021PretrainPA,Radford2021LearningTV,Yao2021CPTCP,Tsimpoukelli2021MultimodalFL}, which enables pre-trained models generalize to downstream tasks with few/zero samples for fine-tuning. Indeed, there already exist many attempts on this topic, for example, \cite{Radford2021LearningTV,Zhou2021LearningTP} make use of crafted templates and learnable continuous representations to reformulate the objective forms of downstream tasks. \cite{Cho2021UnifyingVT,Jin2021AGP,Tsimpoukelli2021MultimodalFL} take account of utilizing an unified text generation framework to uniformly optimize with auto-regressive objective. However, the fixed templates or pre-defined unified generation paradigm may be inadequacy in designing a suitable prompt model owing to the complex semantics in the given questions. To overcome the problem, in this paper we propose an innovative declaration-based prompt model, which exploits question-adaptive declarative sentence as prompt template so that the textual format for VQA task is more consistent with the pre-training phase, diminishing the textual gap between \textit{pre-train} and \textit{fine-tune} stages. 

\section{Methodology}

% 方法部分主要按照总分形式来叙述：
% (1) 先叙述整个方法的框架;
% (2) Preliminary: 主要介绍VQA任务，已经目前VL预训练模型在VQA上的应用策略和方法；
% (3) Declaration-based Prompt Tuning (DPT): 介绍本论文的核心方法部分，按照流程可以分为
%     a. Declarative sentence生成
%     b. 改进文本输入+训练策略；

\begin{figure*}[t]
    \centering
    \includegraphics[width=1.\textwidth]{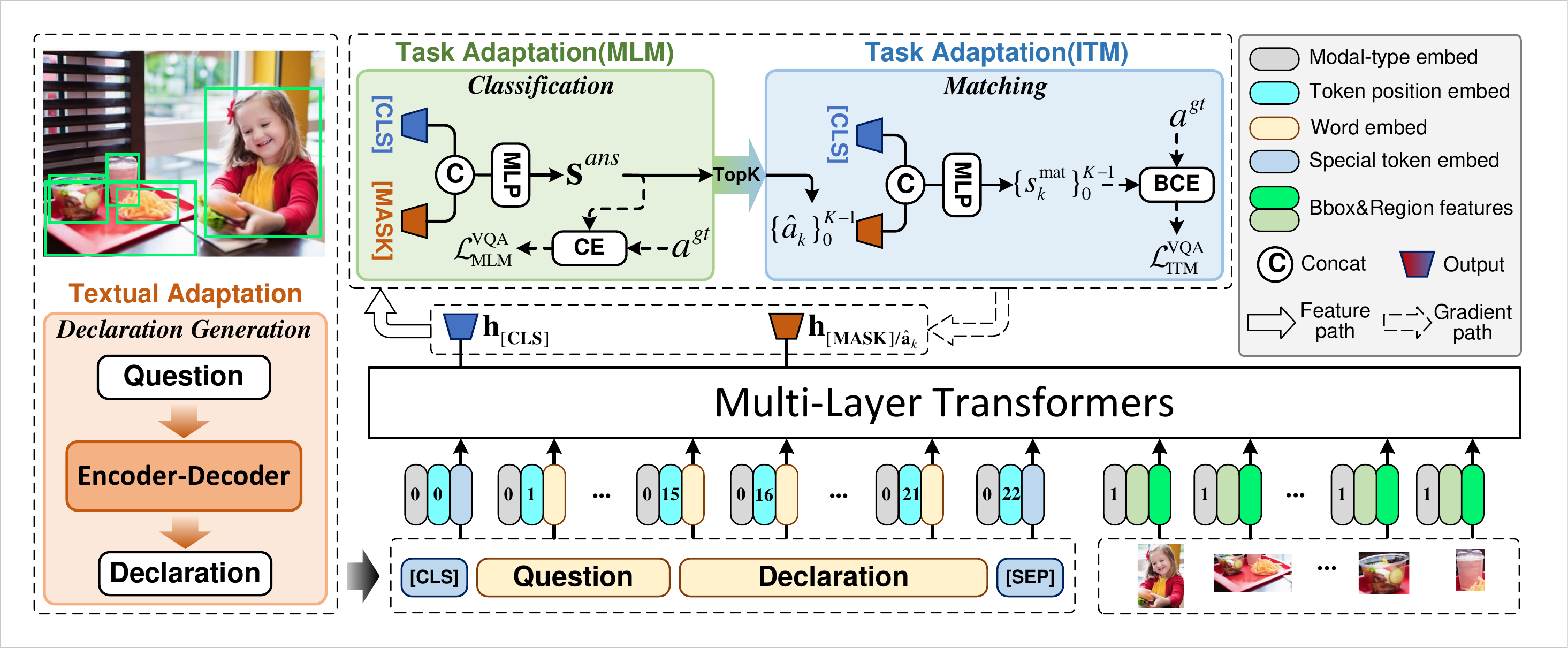} % Reduce the figure size so that it is slightly narrower than the column.
    \caption{The framework of our proposed DPT method. Questions are converted into declarations which are concatenated to form as textual input and fed to the pre-trained VL model along with the regional features. The output [MASK] and [CLS] representations will prompt the model to predict the answer scores or (image-text) matching scores.}
    \label{fig:model}
\end{figure*}

In the following sections, we first present the problem statement of the VQA task (Section \ref{sec:preliminary}). Then, we describe our proposed DPT method (Section \ref{sec:DPT}). The overall framework is depicted in Figure \ref{fig:model}. Specifically, the image and question are converted into the input form and fed to the pre-trained VL model for multi-modal fusion, in which the declaration is typically introduced for prompt tuning. After that, the outputs of the model are exploited to perform the adapted MLM and ITM tasks for model fine-tuning and deciding the answer.

%The overall framework of our \underline{\textbf{D}}eclaration-based \underline{\textbf{P}}rompt \underline{\textbf{T}}uning (DPT) is shown in Figure \ref{fig:model}. In the following sections, we first introduce how the previous pre-trained model solves the VQA task (Section \ref{sec:preliminary}). Then, we describe our proposed DPT method (Section \ref{sec:DPT}), which adapts VQA into pre-training format to facilitate fully-supervised and zero-shot/few-shot learning in the downstream task.

\subsection{Preliminary}
\label{sec:preliminary}

In this paper, we follow the problem definition in \cite{Agrawal2015VQAVQ}, and thus the VQA problem is formulated as a multi-class classification problem. Formally, the VQA task aims to select a correct answer $a$ from a candidate answer set when given an image $I$ and a question $Q$. To this end, we present the classical paradigm for VQA, namely, \textit{pre-training-then-fine-tuning paradigm}.

\noindent\textbf{Pre-training-then-fine-tuning paradigm.} Given a generic architecture, \eg Transformer, the model is first pre-trained on large-scale image-text corpus via manually designed self-supervised tasks, \eg MLM and ITM. To this end, a set of region proposals extracted from the image $I$, $\{\vec{o}_1,\vec{o}_2,...,\vec{o}_n\}$ and word embeddings of the question $Q$, $\{\vec{e}_1,\vec{e}_2,...,\vec{e}_m\}$ are converted to the input format, \ie $\{\vec{e}_{[CLS]},\vec{e}_1,\vec{e}_2,...,\vec{e}_m,\vec{e}_{[SEP]},\vec{o}_1,\vec{o}_2,...,\vec{o}_n\}$, which is fed to the model and fused to produce the hidden representations $\{\vec{h}_i\}_{i=0}^{m+n+2}$, where $\vec{e}_{[CLS]},\vec{e}_{[SEP]}$ are embeddings of special tokens. The model is further optimized using self-supervised objectives. Then, in the fine-tuning stage for VQA task, the output [CLS] is exploited to perform multi-class classification and optimized via cross-entropy loss. This paradigm introduces a brand-new task for fine-tuning, which requires a large amount of labeled data to generalize in downstream task.

\subsection{Declaration-based Prompt Tuning}
\label{sec:DPT}

% 总括该部分的内容
To facilitate the generalization of pre-trained VL models to downstream VQA tasks, we propose a declaration-based prompt tuning (DPT) paradigm that reformulates VQA into pre-training task format. As illustrated in Figure \ref{fig:sample}(b-d), there exist two challenges, \ie different forms of textual input (question \textit{vs.} declaration) and different task objectives (MLM\&ITM \textit{vs.} answer classification). To address these issues, we present (1) \textbf{Textual Adaptation} module to convert questions into their corresponding declarative sentences, and (2) \textbf{Task Adaptation} module to reformulate answer prediction into MLM and ITM tasks. The two adapted tasks are combined to decide the final answer.

\subsubsection{Textual Adaptation via Declaration Generation}
Textual adaptation aims to convert the textual input (\ie questions) into the pre-training form (\ie declarative sentences), \eg the declaration form of ``\textit{What is the red object left of the girl?}'' is ``\textit{A red [MASK] is left of the girl.}''. To this end, we introduce declaration generation which formulates this procedure as a translation problem, where the source and target texts are \textit{question} and corresponding \textit{declaration}, respectively. Formally, we first construct a declaration dataset using the annotations from GQA dataset \cite{Hudson2019GQAAN}, where the ``fullAnswer'' is regarded as the declaration and the short answer word/phrase in ``fullAnswer'' is replaced with a [MASK] token. Then, an encoder-decoder network (T5 \cite{Raffel2020ExploringTL}) is trained on this dataset and optimized using the standard auto-regressive cross-entropy loss. Finally, the model can be used to convert questions into declarative sentences for various VQA datasets, \eg GQA \cite{Hudson2019GQAAN} and VQA \cite{Agrawal2015VQAVQ}. More details are provided in Section \ref{exp:datasets} and Appendix.

\subsubsection{Task Adaptation}
Equipped with declarative sentences, VQA can be reformulated into pre-training task format, \ie MLM and ITM. The adaptations mainly involve two aspects: textual input format and task objectives. Specifically, MLM reserves a [MASK] token in the textual input, and predicts the answer via multi-class classification. ITM replaces [MASK] with the top-k candidate answers predicted from MLM, and predicts the matching scores using binary classification.

\noindent\textbf{Adaptation to MLM task.} To reformulate VQA into MLM task, the question and declaration sentence are concatenated to form as the textual input:
\begin{equation}
    \label{eq:mlm_text_input}
    \mathcal{T}^{MLM}(Q)=\textrm{[CLS]}\ Q\ \textrm{Answer:}\ D\ \textrm{[SEP]}
\end{equation}
where $\mathcal{T}^{MLM}$ represents the conversion function that converts the question $Q$ to the input format. $D$ denotes the declaration sentence. In Equation (\ref{eq:mlm_text_input}), we reserve the question in the textual input, because we find declaration sentence alone drops performance due to the lack of reasoning contexts (refer to Appendix for details). It's worth noting that $D$ reserves a [MASK] token, \eg \textit{a red \textbf{[MASK]} is left of the girl.} In this way, the model is prompted to decide the token to fill in the mask, which exactly indicates the answer word/phrase.

On the basis of the adapted textual input, a pre-trained VL model is exploited to fuse the text and image features, producing a set of hidden representations. The outputs from [CLS] and [MASK] tokens (\ie $\vec{h}_{[CLS]}$ and $\vec{h}_{[MASK]}$) are concatenated to predict the answer:
\begin{gather}\label{eq:mlm_prediction}
    \vec{s}^{ans}=\textrm{MLP}_{MLM}([\vec{h}_{[CLS]};\vec{h}_{[MASK]}]), \\
    p_1(a=a_i|Q,D,I)=\frac{\exp(\vec{s}^{ans}_i)}{\sum_{j=0}^{|\mathcal{C}|}\exp(\vec{s}^{ans}_j)},
\end{gather}
where $\vec{s}^{ans}\in\mathbb{R}^{|\mathcal{C}|}$ denotes the scores over the answer set $\mathcal{C}$. The model is optimized using cross-entropy loss, defined as:
\begin{equation}\label{eq:mlm_loss}
    \mathcal{L}_{MLM}^{VQA}=-\mathbb{E}_{\mathcal{D}}[\log{p_1(a^{gt}|Q,D,I)}],
\end{equation}
where $a^{gt}$ is the ground-truth answer. $\mathcal{D}$ denotes the VQA dataset.

\noindent\textbf{Adaptation to ITM task.} To reformulate VQA into ITM task, the [MASK] token in the declaration sentence $D$ is replaced by the top-k answers $\{\hat{a}_0, \hat{a}_1, ..., \hat{a}_{K-1}\}$ predicted from Equation (\ref{eq:mlm_prediction}), resulting in $K$ candidate declarations:
\begin{equation}\label{eq:itm_text_input}
    \{D^{ans}_0, D^{ans}_1, ..., D^{ans}_{K-1}\}.
\end{equation}

Based on the candidates, the textual input can be formed via concatenation of the question $Q$ and the declaration sentence $D^{ans}_k$, defined as follows:
\begin{equation}
    \mathcal{T}^{ITM}(Q)=\textrm{[CLS]}\ Q\ \textrm{Answer:}\ D^{ans}_k\ \textrm{[SEP]}
\end{equation}
where $\mathcal{T}^{ITM}$ represents the conversion function. $D^{ans}_k$ denotes the declaration sentence, in which the [MASK] token is replaced by the $k$-th candidate answer $\hat{a}_{k}$, \eg \textit{a red \textbf{tray/food/cloth} is left of the girl.}

In this way, pre-trained VL models are prompted to determine whether the image-text is matched. To achieve this, the image and textual inputs are fed to the VL model, and the outputs from [CLS] and answer token (\ie $\vec{h}_{[CLS]}$ and $\vec{h}_{\hat{a}_k}$) are concatenated to predict the matching score:
\begin{gather}\label{eq:itm_prediction}
s^{mat}_{k}=\textrm{MLP}_{ITM}([\vec{h}_{[CLS]};\vec{h}_{\hat{a}_k}]), \\
p_2(a=\hat{a}_k|Q,D^{ans}_k,I)=\textrm{sigmoid}(s^{mat}_{k}),
\end{gather}
where $s^{mat}_k$ denotes the matching score of the image and the $k$-th candidate answer. Intuitively, the image-text pair with ground-truth answer should have higher matching score. Therefore, the model is optimized using binary cross-entropy loss, defined as follows:
\begin{gather}\label{eq:itm_loss}
    y_k = \mathbb{I}[\hat{a}_k=a^{gt}], \\
    \resizebox{.89\linewidth}{!}{$
    \displaystyle
    \mathcal{L}_{ITM}^{VQA}=-\mathbb{E}_{\mathcal{D}}\frac{1}{K}\sum_{k=0}^{K-1}[y_k\log{p_2(\hat{a}_k)} + (1 - y_k)\log{(1 - p_2(\hat{a}_k))}],
$}
\end{gather}
where $\mathbb{I}[x]: X\rightarrow\{0,1\}$ denotes the indicator function, which takes value 1 if $x$ is positive and zero otherwise.

\noindent\textbf{Training and inference.} On the top of task adaptation, VQA has been reformulated into MLM and ITM problems. During training, we integrate the loss terms from Eq. (\ref{eq:mlm_loss}) and (\ref{eq:itm_loss}) to fine-tune VL models. The total loss of DPT is defined as:
\begin{equation}
    \mathcal{L}_{DPT}=\mathcal{L}_{MLM}^{VQA} + \mathcal{L}_{ITM}^{VQA},
\end{equation}
During inference, the normalized scores predicted by MLM and ITM are combined via simple summation, and the answer $\hat{a}$ with the highest score is chosen as the final prediction result, defined as follows:
\begin{gather}
    \hat{a}=\mathop{\textrm{argmax}}\limits_{\hat{a}\in{\{\hat{a}_i\}_{0}^{K-1}}}(p_1(\hat{a}) + p_2(\hat{a})),
\end{gather}

\noindent\textbf{Zero-shot and few-shot learning.} Equipped with DPT, previous pre-trained VL models can also be easily transformed for zero-shot or few-shot learning based VQA tasks, only if reformulating Equation (\ref{eq:mlm_prediction}) and (\ref{eq:itm_prediction}) into the same form as the one in pre-trained phrase, and is initialized with the pre-trained weights, which can be rewritten as follows,  
%perform zero-shot or few-shot learning for VQA task. The key point is to reformulate Equation (\ref{eq:mlm_prediction}) and (\ref{eq:itm_prediction}) into the form that is exactly same as pre-trained phase, and initialize the parameters using the pre-trained weights, which can be rewritten as follows,
\begin{gather}
    \vec{s}^{ans}=\textrm{MLP}^{pt}_{MLM}(\vec{h}_{[MASK]}), \\
    s^{mat}_{k}=\textrm{MLP}^{pt}_{ITM}(\vec{h}_{[CLS]}),
\end{gather}
where $\textrm{MLP}^{pt}_{*}$ denotes the MLP layer initialized with pre-trained weights. Since the number of answers is less than that of vocabulary tokens, only the weights corresponding to answer words are taken to initialize $\textrm{MLP}^{pt}_{MLM}$. 

% 实验部分主要分为几小节：
% (1) Implementation Details
%     a. Datasets
%     b. Declaration generation training
%     c. VQA fine-tuning
%     d. Model evaluation and comparison
% (2) Model Performance
%     a. Comparison with the state of the art.
%     b. ...
% (3) Visualization
%     a. Samples with different predictions

\section{Experiments}

\subsection{Implementation Details}
\paratitle{Datasets.}\label{exp:datasets} GQA \cite{Hudson2019GQAAN} and VQA v2.0 \cite{Agrawal2015VQAVQ} are used to build declaration generation dataset and evaluate our proposed methods on VQA task. More details are provided in the Appendix.

\paratitle{Model training.}\label{exp:training} T5-small \cite{Raffel2020ExploringTL} is chosen for declaration generation. As for VQA, VinVL \cite{Zhang2021VinVLRV} is selected as our base architecture. Our proposed DPT is applied to VinVL via textual and task adaptation. The model is fine-tuned using the adapted task objectives, resulting in two variants regarding the tasks for training, \ie DPT(MLM) and DPT(MLM\&ITM). The number of answers used for ITM $K$ is set to 8. For fair comparison, we follow the same training settings as reported in the previous works in the following experiments. The details of hyper-parameters are reported in Appendix.

\subsection{Experimental Results}

% Table generated by Excel2LaTeX from sheet 'Presentation'
\begin{table}
    \centering
      \begin{tabular}{lccc}
      \toprule
      \multirow{2}[2]{*}{Method} & \multirow{2}[2]{*}{Pre-trained} & \multicolumn{2}{c}{Accuracy (\%)} \\
            &       & Test-dev & Test-std \\
      \midrule

      MMN [\citeyear{Chen2021MetaMN}]   & \multirow{2}[2]{*}{\XSolidBrush} & - & 60.83  \\
      NSM [\citeyear{Hudson2019LearningBA}]  &       & - & 63.17  \\
      \midrule
      LXMERT [\citeyear{Tan2019LXMERTLC}] & \multirow{7}[2]{*}{\Checkmark} & 60.00  & 60.33  \\
      VILLA [\citeyear{Gan2020LargeScaleAT}] &       & 60.98  & 61.12  \\
      OSCAR [\citeyear{Li2020OscarOA}] &       & 61.58  & 61.62  \\
      VL-T5 [\citeyear{Cho2021UnifyingVT}] &       & - & 60.80  \\
      MDETR [\citeyear{Kamath2021MDETRM}] &       & 62.95  & 62.45  \\
      VinVL$_{bal}^{\dag}$ [\citeyear{Zhang2021VinVLRV}] &     &  60.76 & 60.89 \\
      VinVL [\citeyear{Zhang2021VinVLRV}] &       & 65.05 & 64.65 \\
      \midrule
      DPT$_{bal}$ &    \multirow{2}[2]{*}{\Checkmark}   & 63.55 & 63.57 \\
      DPT   &   & \textbf{65.20} & \textbf{64.92} \\
      \bottomrule
      \end{tabular}%
      \caption{Accuracy comparisons over the GQA dataset. `-' and '`$\dag$' denote the numbers are not available and our implementation, respectively. ${bal}$ denotes the model trained on the \textit{balanced} split.}
    \label{tab:gqa_test}%
\end{table}%

For online evaluation of GQA dataset, we compare our method with the state-of-the-art models, including non-pretrained models \ie MMN \cite{Chen2021MetaMN}, NSM \cite{Hudson2019LearningBA}, and pre-trained VL models \ie LXMERT \cite{Tan2019LXMERTLC}, VILLA \cite{Gan2020LargeScaleAT}, OSCAR \cite{Li2020OscarOA}, VinVL \cite{Zhang2021VinVLRV}, MDETR \cite{Kamath2021MDETRM}, VL-T5 \cite{Cho2021UnifyingVT}. The results are reported in Table \ref{tab:gqa_test}. When only exploiting \textit{balanced} split for training, our method achieves 63.55\% and 63.57\% overall accuracy on test-dev and test-std, respectively, outperforming the state-of-the-art non-pretrained/pre-trained models. Specifically, our method (DPT$_{bal}$) surpasses the fine-tuned counterpart (VinVL$_{bal}$) by a significant margin of 2.68\% on test-std. When using \textit{all} split to bootstrap our model similar to \cite{Chen2021MetaMN,Zhang2021VinVLRV}, our method (DPT) still ranks the top regarding overall accuracy, and outperforms the counterpart (VinVL) by 0.27\% on test-std. Among the compared models, MMN and NSM also achieve competitive results even if no pre-training is performed, which is thanks to the usage of deliberately generated scene graphs or supervision of execution programs.

% (1) different prompt strategies
% (2) generalizability over different datasets
% (3) generalizability over different VL models
% (4) zero-shot & few-shot
\subsection{Ablation Study}

For a deeper understanding of DPT, we further conduct the ablation studies on the local validation split of GQA and VQA v2.0 datasets (\emph{textdev} on GQA and \emph{val} on VQA v2.0).

% Table generated by Excel2LaTeX from sheet 'Presentation'
\begin{table}
  \centering
    \begin{tabular}{lcccc}
    \toprule
    \multirow{2}[2]{*}{Prompt} & \multirow{2}[2]{*}{Output} & \multirow{2}[2]{*}{Task} & \multicolumn{2}{c}{Accuracy (\%)} \\
          &       &       & GQA   & VQA v2 \\
    \midrule
    Baseline & [C]   & Baseline & 60.26  & 74.05  \\
    Mask  & [C]\&[M] & MLM   & 60.88  & 74.30  \\
    Dynamic & [C]\&[M] & MLM   & 62.09  & 74.39  \\
    \midrule
    \multirow{3}[2]{*}{Declaration} & [M]   & MLM   & 60.03  & 73.90  \\
          & [C]\&[M] & MLM   & 62.71  & 74.39  \\
          & [C]\&[M] & MLM\&ITM & \textbf{63.13 } & \textbf{74.50 } \\
    \bottomrule
    \end{tabular}%
    \caption{Effectiveness validation of declarative sentences for prompt tuning on GQA and VQA v2.0 datasets. \textit{Output} and \textit{Task} denote the outputs for prediction and the adapted tasks for fine-tuning, respectively. [C] and [M] are abbreviated as [CLS] and [MASK].}
  \label{tab:ablation_prompt}%
\end{table}%
  
\paratitle{Different prompts.} To illustrate the effectiveness of declarative sentences for prompt tuning, several prompt variants are proposed for comparison in Table \ref{tab:ablation_prompt}, defined as follows:
\begin{itemize}
    \item \textbf{Baseline}: Vanilla fine-tuning VinVL \cite{Zhang2021VinVLRV} without prompt.
    \item \textbf{Mask}: ``Answer: [MASK]''.
    \item \textbf{Dynamic}: ``Answer: [V1] [V2] ... [V16] [MASK]''.
    \item \textbf{Declaration (Ours)}: ``Answer: $D$''.
\end{itemize}
where `[V1]'-`[V16]' denote the learnable tokens which are jointly trained during fine-tuning. As Table \ref{tab:ablation_prompt} shows, on GQA dataset, our proposed declaration-based prompt is more effective than manually designed templates (\ie Mask and Dynamic). For example, DPT with MLM task (row 5) surpasses the Mask and Dynamic with 1.83\% and 0.62\%, respectively. Equipped with both MLM and ITM tasks, our full model (row 6) surpasses Baseline by 2.87\%. To measure the confidence of the results, we have performed additional 3 runs for our best-performing model on GQA and VQA v2.0 datasets, getting standard deviations of 0.10\% and 0.06\%, respectively.

% Table generated by Excel2LaTeX from sheet 'Presentation'
\begin{table}
    \centering
      \begin{tabular}{lccc}
      \toprule
      Model & Task  & Accuracy (\%) & $\Delta$ (\%) \\
      \midrule
      \multirow{3}[2]{*}{VinVL[\citeyear{Zhang2021VinVLRV}]} & Baseline & 74.05  & 0.00  \\
            & MLM   & 74.39  & 0.34  \\
            & MLM\&ITM & \textbf{74.50 } & \textbf{0.45 } \\
      \midrule
      \multirow{3}[2]{*}{ViLT[\citeyear{Kim2021ViLTVT}]} & Baseline & 70.71  & 0.00  \\
            & MLM   & 71.01  & 0.30  \\
            & MLM\&ITM & \textbf{71.17 } & \textbf{0.46 } \\
      \midrule
      \multirow{3}[2]{*}{UNITER[\citeyear{Chen2020UNITERUI}]} & Baseline & 67.72  & 0.00  \\
            & MLM   & 68.69  & 0.97  \\
            & MLM\&ITM & \textbf{68.73 } & \textbf{1.01 } \\
      \bottomrule
      \end{tabular}%
      \caption{Effectiveness validation of DPT over different pre-trained VL models on VQA v2.0 datasets. $\Delta (\%)$ denotes the absolute accuracy improvement margin compared with baseline.}
    \label{tab:ablation_vl_models}%
\end{table}%

\paratitle{Generalizability over different datasets.} Table \ref{tab:ablation_prompt} shows the ablation results on VQA v2.0 with respect to different prompts. Consistent with the results on GQA, our proposed DPT surpasses the fine-tuning using fixed templates, \ie Mask or Dynamic. Specifically, our model with DPT outperforms Baseline by 0.45\%. The difference in accuracy gain between GQA and VQA (2.87\% \emph{vs.} 0.45\%) is mainly due to the question complexity and the quality of the generated declaration sentences (refer to Appendix for details).

\paratitle{Generalizability over different VL models.} To illustrate the generalizability of our proposed method over different pre-trained VL models, we apply our DPT to the recently proposed VL models that have been pre-trained via MLM and ITM tasks, \eg UNITER \cite{Chen2020UNITERUI} and ViLT \cite{Kim2021ViLTVT}. As shown in Table \ref{tab:ablation_vl_models}, for all the three baselines, equipped with our DPT method, a consistent performance improvement (0.64\% on average) can be observed. For example, ViLT+DPT and UNITER+DPT achieve absolute performance gains of 0.46\% and 1.01\% compared with the fine-tuning counterparts, respectively. 

\begin{figure}[t]
    \centering
    \includegraphics[width=1.\columnwidth]{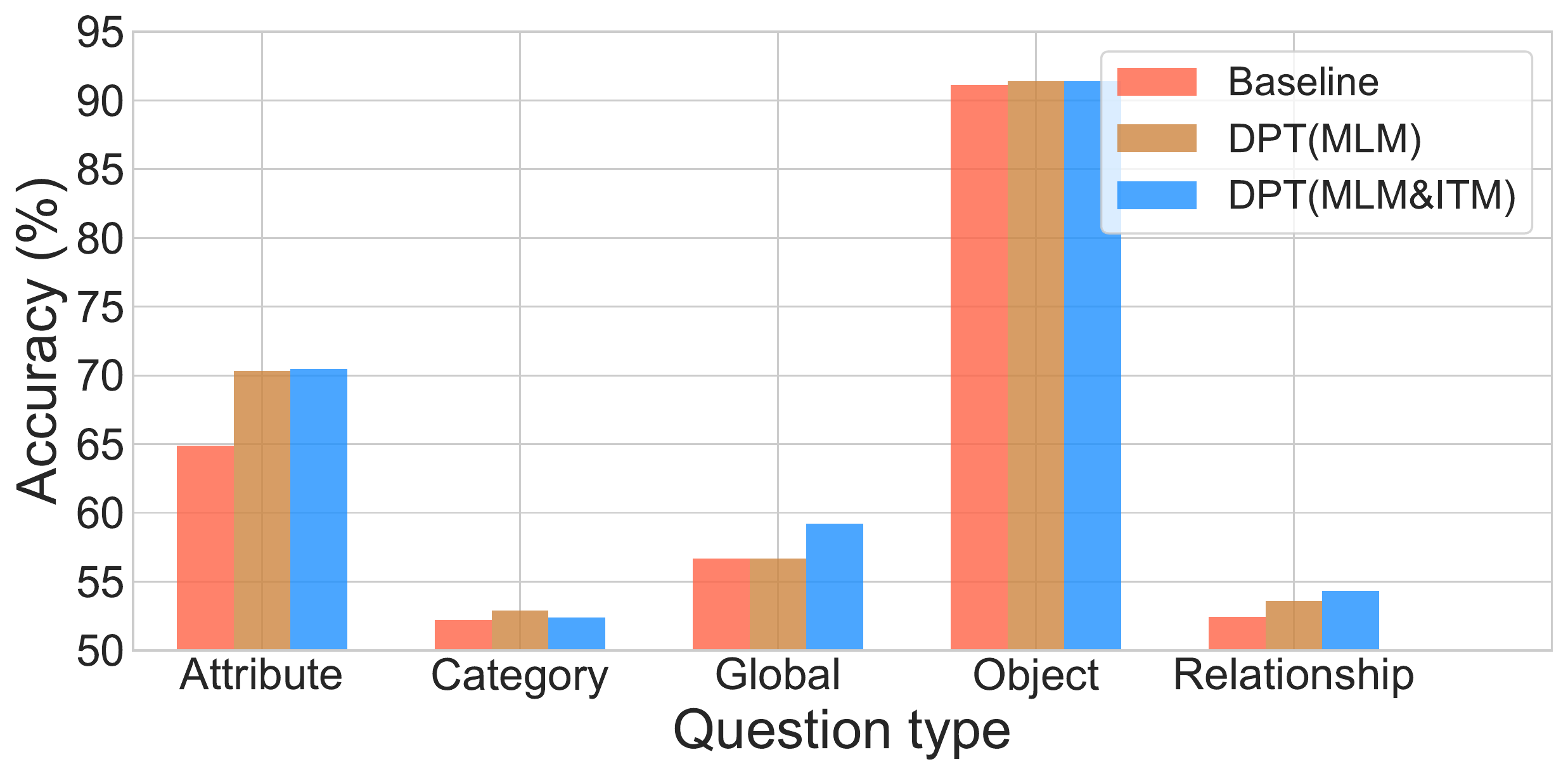} % Reduce the figure size so that it is slightly narrower than the column. Don't use precise values for figure width.This setup will avoid overfull boxes.
    \caption{Accuracy breakdown over question semantic types on the GQA dataset. }
\label{fig:accuracy_over_types}
\end{figure}

\paratitle{Accuracy over different question types.} Figure \ref{fig:accuracy_over_types} shows the accuracy breakdown on different question semantic types. It can be observed that the adapted MLM task achieves large accuracy improvement in attribute questions against Baseline (70.46\% \textit{vs.} 64.87\%). This shows the strength of declaration-based prompt in capturing the object attributes. Moreover, the adapted ITM task brings more performance improvement in global questions (59.24\% \textit{vs.} 56.69\%), indicating its superior ability in the understanding of global semantics.

\begin{figure}[t]
    \centering
    \includegraphics[width=1.\columnwidth]{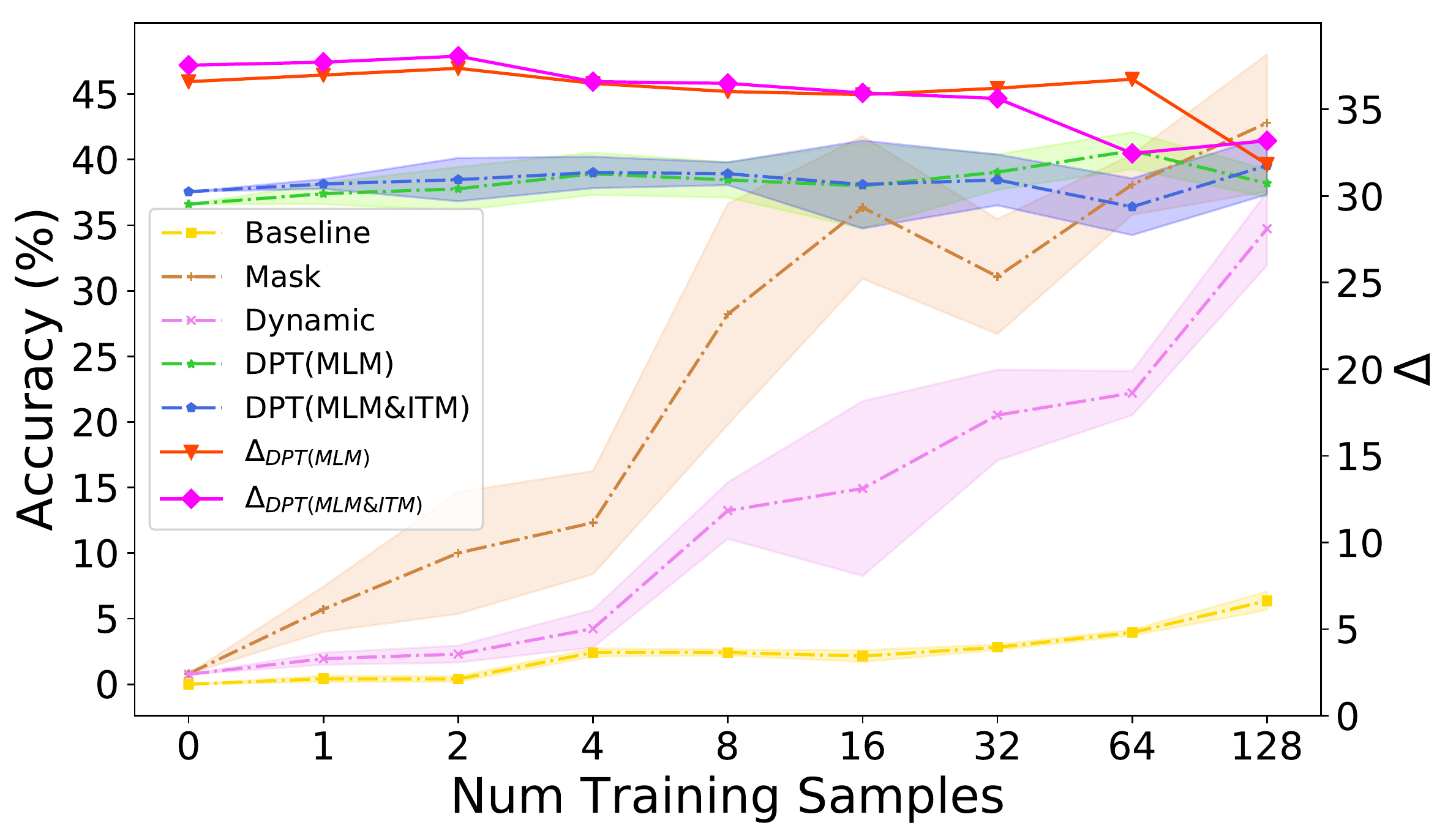} % Reduce the figure size so that it is slightly narrower than the column. Don't use precise values for figure width.This setup will avoid overfull boxes.
    \caption{Testdev accuracy in zero-shot and few-shot settings on GQA dataset. The mean and standard deviation are reported over 5 random splits. $\Delta$ indicates the absolute accuracy improvement compared to Baseline.}
\label{fig:accuracy_few_shot_wo_yn}
\end{figure}

\subsection{Zero-shot and Few-shot Results}\label{sec:zero_few_shot}

Figure \ref{fig:accuracy_few_shot_wo_yn} shows the accuracy in zero-shot and few-shot settings on GQA dataset. We remove \textit{yes/no} questions in the sampled splits in advance since the large proportion of \textit{yes/no} questions (18.81\% and 17.47\% questions have \textit{yes} and \textit{no} answers, respectively) will cause large variance ($\sim$8\%) in Baseline evaluation. As shown in Figure \ref{fig:accuracy_few_shot_wo_yn}, it can be observed that DPT outperforms the vanilla fine-tuning counterparts and other prompt variants (\ie Mask and Dynamic) by a significant margin. For example, with no samples for training, our DPT achieves a strong accuracy of 36.6\% while the fine-tuning counterpart can not predict correct answers due to random guessing. When provided 1$\sim$128 samples, our DPT method achieves 31.8\%$\sim$37.4\% absolute accuracy improvement compared to Baseline.

\begin{figure}[t]
    \centering
    \includegraphics[width=1.\columnwidth]{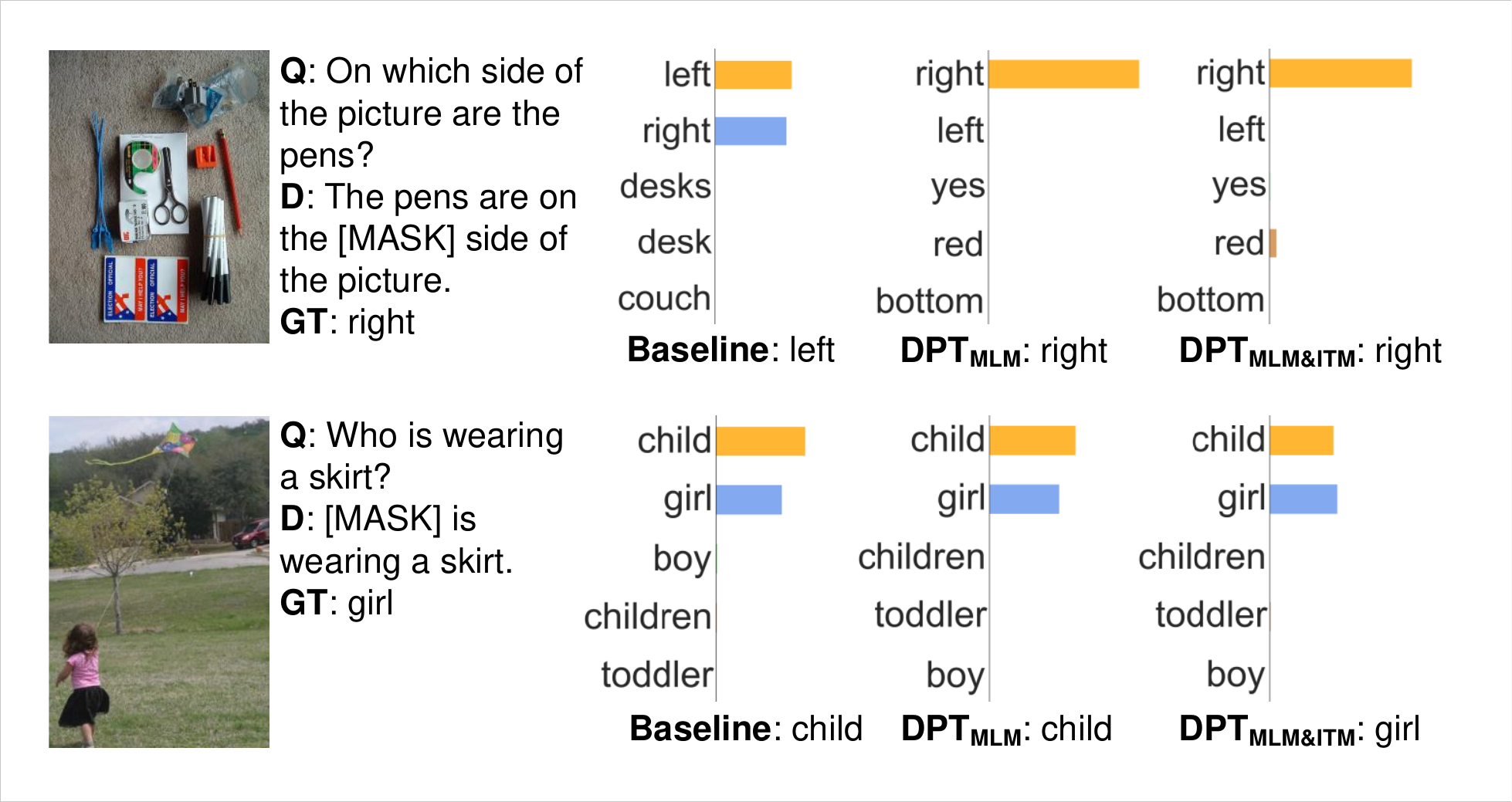} % Reduce the figure size so that it is slightly narrower than the column. Don't use precise values for figure width.This setup will avoid overfull boxes.
    \caption{Visualization of the predictions from Baseline and our proposed DPT method. \textbf{Q}, \textbf{D} and \textbf{GT} denote question, generated declarative sentence and ground-truth answer, respectively.}
\label{fig:cases}
\end{figure}

\subsection{Case study}

In Figure \ref{fig:cases}, we visualize two successful cases from our proposed DPT method. Regarding the first case, the baseline yields almost the same probabilities for `left' and `right', indicates its weakness in solving such direction-related questions. In contrast, equipped with the ability of masked language model, our DPT confidently predicts the correct answer `right'. As for the second case, the baseline model and $\textrm{DPT}_{\textrm{MLM}}$ both wrongly predict the answer `child' mainly attribute to `child' being a more frequent object that occurs in the train set. Besides, `child' is a hypernym of `girl' and `boy', making it a universal answer to many questions. On the contrary, DPT with the adapted ITM task takes account of the semantics of answers, and gives a higher score to the answer `girl', leading to the correct answer.

\section{Conclusion}

We propose to reformulate the VQA task into masked language model (MLM) and image-text matching (ITM) problems, maximally mitigating the gap between vision-language (VL) pre-training and fine-tuning stages. To achieve this, we first convert questions into declarative sentences with reserved [MASK] or candidate answers, mitigating the discrepancies regarding the textual input. Then, VQA problem is reformulated into pre-training format via task adaptation, which solves VQA in the manner of MLM and ITM tasks. Extensive experiments on two benchmarks validate the effectiveness and generalizability of our proposed DPT paradigm over different pre-trained VL models in both fully-supervised and zero-shot/few-shot settings.

\section{Acknowledgements}
This work was supported in part by the National Natural Science Foundation of China under Grant No.61602197, Grant No.L1924068, Grant No.61772076, in part by CCF-AFSG Research Fund under Grant No.RF20210005, in part by the fund of Joint Laboratory of HUST and Pingan Property \& Casualty Research (HPL), and in part by the National Research Foundation (NRF) of Singapore under its AI Singapore Programme (AISG Award No: AISG-GC-2019-003). Any opinions, findings and conclusions or recommendations expressed in this material are those of the authors and do not reflect the views of National Research Foundation, Singapore. The authors would also like to thank the anonymous reviewers for their comments on improving the quality of this paper.

%% The file named.bst is a bibliography style file for BibTeX 0.99c
\bibliographystyle{main}
\bibliography{main}

\clearpage
\appendix
\section*{Appendix}

% (1) Datasets
% (2) Model training and evaluation
% (3) Baselines
% (4) Additional experiments
% (5) Case study

\section{Datasets}\label{sec:appendix_dataset}
\paratitle{VQA v2.0.} VQA v2.0 \cite{Agrawal2015VQAVQ} is the most commonly used VQA benchmark. It contains real images and annotated question-answer pairs. Each image has an average of 5 questions. Each question has 10 answers annotated by different annotators, and the most frequent answer is regarded as the ground-truth answer. The dataset is split into \textit{train}, \textit{val}, and \textit{test} sets, statistically detailed in Table \ref{tab:dataset_VQA}. The evaluation metric (\ie accuracy) on this dataset is robust to inter-human variability, calculated as follows:
\begin{equation}
    \textrm{Acc}(\textrm{\textbf{ans}})=\textrm{min}\Big{\{} \frac{\#\textrm{humans that said \textbf{ans}}}{3}, 1\Big{\}}
\end{equation}

% Table generated by Excel2LaTeX from sheet 'VQA'
% \renewcommand{\arrarstretch}{1.3}
\begin{table}[b]
  \centering
    \begin{tabular}{|c|c|c|c|}
    \hline
    Split & \#Images & \#Questions & \#Answers \\
    \hline
    Train & 82,783 & 443,757 & 4,437,570 \\
    Val   & 40,504 & 214,354 & 2,143,540 \\
    Test  & 81,434 & 447,793 & - \\
    \hline
    All   & 204,721 & 1,105,904 & - \\
    \hline
    \end{tabular}%
    \caption{Statistics of samples in VQA-v2 dataset}
  \label{tab:dataset_VQA}%
\end{table}%

\paratitle{GQA.} GQA \cite{Hudson2019GQAAN} is a VQA dataset that characterizes in compositional question answering and visual reasoning about real-world images. With the help of the scene graph annotations from Visual Genome \cite{Krishna2016VisualGC}, GQA is able to maximally mitigate the language priors that exist widely in previous VQA datasets. Additionally, the questions are generated via the engine that operates over 524 patterns, spanning 117 question groups. Therefore, it requires more complicated reasoning skills to answer the questions. GQA consists of two splits, \ie \textit{all} split that contains 22M QA pairs, and \textit{balanced} split that consists of 1.7M QA pairs with resampled question-answer distribution. The dataset is split into 70\% train, 10\% validation, 10\% test and 10\% challenge. The statistics of \textit{balanced} split are detailed in Table \ref{tab:dataset_GQA}.

% Table generated by Excel2LaTeX from sheet 'VQA'
\begin{table}[t]
  \centering
    \begin{tabular}{|c|c|c|c|}
    \hline
    Split & \#Images & \#Questions & \#Vocab \\
    \hline
    Train & 72,140 & 943,000 & \multirow{4}[2]{*}{3,097} \\
    Val   & 10,234 & 132,062 &  \\
    Test-dev & 398   & 12,578 &  \\
    Test  & 2,987 & 95,336 &  \\
    \hline
    All   & 85,759 & 1,182,976 & 3,097 \\
    \hline
    \end{tabular}%
    \caption{Statistics of balance-split in GQA dataset}
  \label{tab:dataset_GQA}%
\end{table}%

\paratitle{Declaration dataset.} As introduced in Section \ref{exp:datasets}, the declaration dataset is generated from the annotations from GQA dataset. Specifically, each annotation in GQA training set contains three keys, \ie \textit{question}, \textit{answer}, and \textit{fullAnswer}, illustrated in Table \ref{tab:gqa_sample}.
% Table generated by Excel2LaTeX from sheet 'Examples'
\begin{table}[t]
    \centering
      \begin{tabular}{|cc|}
      \toprule
      questionId & 201640614 \\
      \midrule
      question & Who is wearing the dress? \\
      answer & woman \\
      fullAnswer & The woman is wearing a dress. \\
      \bottomrule
      \end{tabular}%
      \caption{An annotation example of GQA dataset. Only part of the annotations is listed for illustration.}
    \label{tab:gqa_sample}%
\end{table}%
It can be observed that `answer' is usually a word or phrase while the `fullAnswer' is a complete declarative sentence. According to statistics, we find that most short `answer's are included in the `fullAnswer's, which inspires us that instead of choosing one from a candidate answer set, we can extract a word/phrase as the answer from the complete declarative sentence in the form of cloze-filling. Formally, we replace the short `answer' included in the `fullAnswer' with a [MASK] token, resulting in a declarative sentence that can be used for answer clozing, illustrated in Table \ref{tab:qtd_sample}.
% Table generated by Excel2LaTeX from sheet 'Examples'
\begin{table}[t]
    \centering
      \begin{tabular}{|cc|}
      \toprule
      questionId & 201640614 \\
      \midrule
      question & Who is wearing the dress? \\
      declaration & The [MASK] is wearing a dress. \\
      \bottomrule
      \end{tabular}%
      \caption{An example of question and the corresponding declaration.}
    \label{tab:qtd_sample}%
  \end{table}%

In this way, we are able to convert the questions into declaration format from the \textit{all} split of GQA dataset, resulting in 726k and 181k question-declaration pairs for training and validation, respectively. Several examples from our proposed declaration dataset are shown in Table \ref{tab:qtd_examples}.
% Table generated by Excel2LaTeX from sheet 'Examples'
\begin{table}[t]
    \centering
      \begin{tabular}{|cl|}
      \toprule
      \textbf{Type} & \textbf{Content} \\
      \midrule
      Q     & Are there any black gloves or hats? \\
      D     & [MASK] \\
      \midrule
      Q     & What is the vehicle that is to the right of the man? \\
      D     & the vehicle is a [MASK]. \\
      \midrule
      Q     & On which side of the image is the woman? \\
      D     & the woman is on the [MASK] of the image. \\
      \midrule
      Q     & Who is swinging the bat? \\
      D     & the [MASK] is swinging the bat. \\
      \midrule
      Q     & What color are the gloves? \\
      D     & the gloves are [MASK]. \\
      \bottomrule
      \end{tabular}%
      \caption{Five samples from the declaration dataset. Q and D indicate question and declaration, respectively.}
    \label{tab:qtd_examples}%
\end{table}%

\section{Model Training and Evaluation}
\label{sec:train_val_detail}

\paratitle{Declaration generation training.} We use T5-small \cite{Raffel2020ExploringTL} as our declaration generation model due to its transferable abilities in unified text-to-text generation. Specifically, the generation process is regarded as a \textit{translation} problem, in which \textit{question} and \textit{declaration} represent the source and target texts with the same vocabulary, respectively. The model is fine-tuned on the training set and evaluated on the validation set. The parameters are trained via Adam optimizer with learning rate of $5e^{-5}$ and batch size of $4$ for 480k steps in total. The fine-tuned model can be used to convert questions for GQA and VQA v2.0 datasets. 

\paratitle{GQA training and evaluation.} Since GQA consists of two splits, \ie \textit{balanced} and \textit{all}, we evaluate our proposed methods on both splits. With regard to the \textit{balanced} split, the model is trained on the concatenation of \textit{train+val} and evaluated on the \textit{testdev}. As for \textit{all} split, we follow VinVL \cite{Zhang2021VinVLRV} to set up the fine-tuning procedure, which fine-tunes the model on \textit{all} split for 5 epochs and then fine-tunes on the \textit{balanced} split for 2 epochs. We evaluate our methods on both splits in the online evaluation. As for local ablation study, only \textit{balanced} split is used.

\paratitle{VQA v2.0 training and evaluation.} VQA v2.0 is typically designed to assess the generalizability of our proposed methods. Therefore, the model is trained on the \textit{train} split and evaluated on local \textit{valid} split. As for model training, we follow the same settings as the previous works (\ie VinVL \cite{Zhang2021VinVLRV}, ViLT \cite{Kim2021ViLTVT} and UNITER \cite{Chen2020UNITERUI}), on which our DPT is implemented based.

\section{Additional Experimental Results}

\subsection{Declaration Generation}

The commonly used metric in machine translation, \ie Bilingual Evaluation Understudy (BLEU) is adopted to evaluate the translation quality of the generated declarative sentences. Specifically, BLEU calculates scores based on the overlap of words in the reference text and the generated text. The higher score of BLEU denotes the better generation quality. We evaluate the best T5 model on the 181k validation set, getting BLEU score of 0.97, indicating that the fine-tuned T5 model is able to generate declarative sentences of very high quality.

\subsection{Number of Candidate Answers}

Table \ref{tab:ablation_k} shows the ablation results over the number of candidate answers for image-text matching. It can be observed that the best performance 63.13\% is achieved when $K=8$. Additionally, computational complexity will increase dramatically when $K$ grows larger, while the accuracy does not change much. Therefore, we set $K=8$ throughout the experiments.

% Table generated by Excel2LaTeX from sheet 'Presentation'
\begin{table}[t]
  \centering
    \begin{tabular}{cc}
    \toprule
    K     & Accuracy (\%) \\
    \midrule
    0     & 62.79  \\
    1     & 62.79  \\
    2     & 62.90  \\
    4     & 63.06  \\
    8     & \textbf{63.13}  \\
    16    & 63.12  \\
    \bottomrule
    \end{tabular}%
    \caption{Ablation experiments on the GQA balanced dataset. $K$ stands for the number of candidate answers chosen for image-text matching.}
  \label{tab:ablation_k}%
\end{table}%

\subsection{Accuracy Breakdown Analysis}
\label{sec:accuracy_over_length}

\begin{figure}[t]
  \centering
  \includegraphics[width=1.\columnwidth]{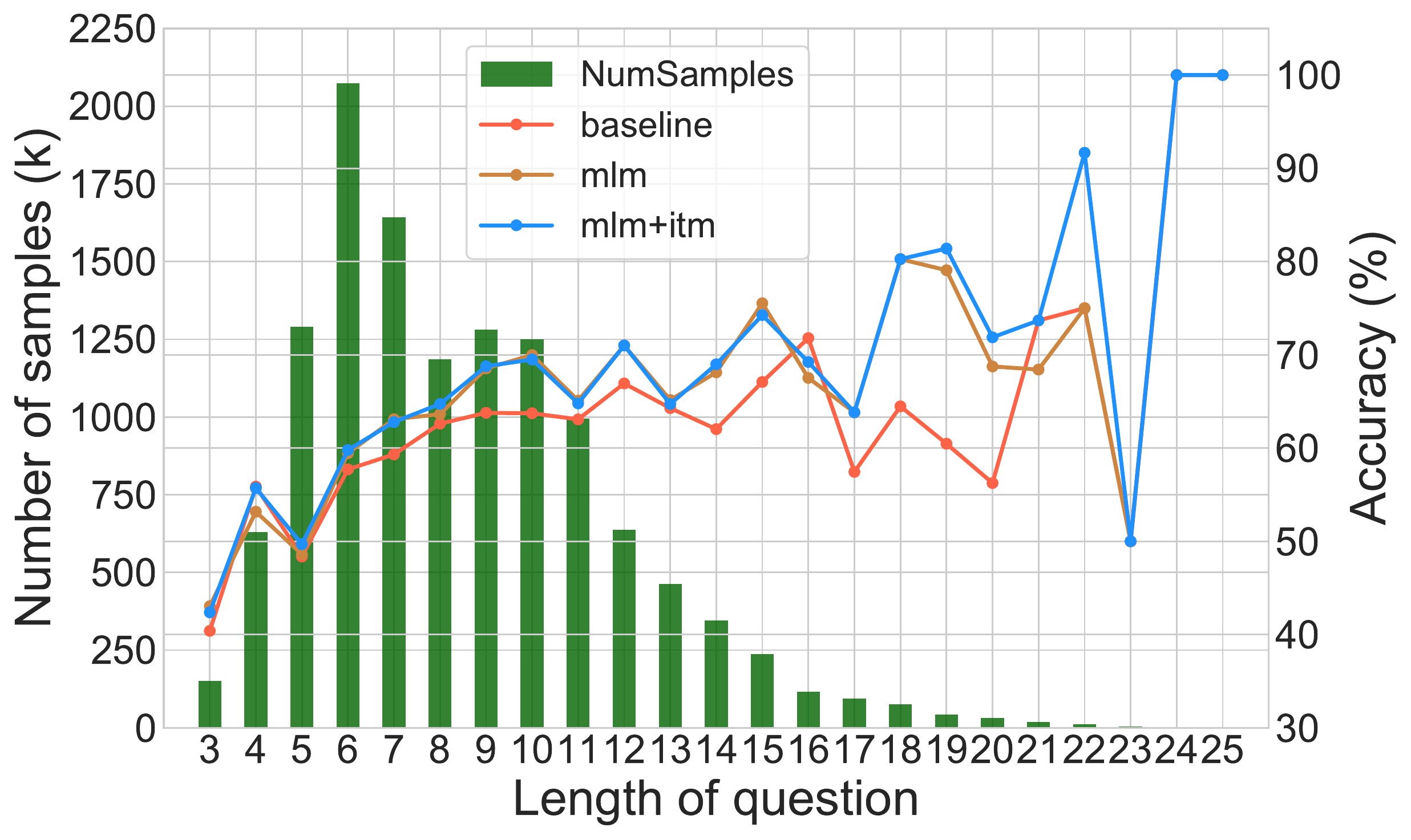} % Reduce the figure size so that it is slightly narrower than the column. Don't use precise values for figure width.This setup will avoid overfull boxes.
  \caption{Accuracy breakdown over question length on GQA dataset. }
\label{fig:accuracy_over_length}
\end{figure}

Figure \ref{fig:accuracy_over_length} shows the accuracy breakdown over the question length on \textit{testdev} of GQA dataset. Generally, the question length reflects the reasoning complexity to a certain extent. As Figure \ref{fig:accuracy_over_length} shows, it can be observed that our DPT obtains more accuracy improvement on the question with length of 14$\sim$20, while achieves similar accuracy on the short questions (with length of 3$\sim$13), demonstrating the effectiveness of our proposed DPT over long questions.

\section{Case Study}

\subsection{Declaration Generation}
\label{sec:declaration_generation}

To perform predictions on the test split of GQA or VQA v2.0 datasets, the fine-tuned T5 model is exploited to convert questions into declarative sentences. In Table \ref{tab:qtd_predictions}, we visualize several declarative sentences from GQA dataset, which are generated by the fine-tuned T5 model. It can be observed that [MASK] token is placed at the proper position that can prompt the model to predict the answer. However, we find that some generated sentences may not contain complete semantic information of the question. For example, the sample-3 refers to \textit{vehicle \textbf{in front of the flag}}, while only \textit{vehicle} is generated in the declarative sentence. This may cause ambiguity in visual reasoning. Therefore, we preserve the original question in the textual input of VL models, maximally reserving the semantics of questions.

% Table generated by Excel2LaTeX from sheet 'Examples'
\begin{table}[t]
  \centering
    \begin{tabular}{|cll|}
    \toprule
    \textbf{Id} & \textbf{Type} & \textbf{Content} \\
    \midrule
    \multirow{2}[2]{*}{0} & Q     & Which color is the shirt? \\
          & D     & the shirt is [MASK]. \\
    \midrule
    \multirow{2}[2]{*}{1} & Q     & What is beneath the microwave? \\
          & D     & the [MASK] is beneath the microwave. \\
    \midrule
    \multirow{2}[2]{*}{2} & Q     & Is this a bed or a cabinet? \\
          & D     & this is a [MASK]. \\
    \midrule
    \multirow{2}[2]{*}{3} & Q     & Which kind of vehicle is in front of the flag? \\
          & D     & the vehicle is a [MASK]. \\
    \midrule
    \multirow{2}[2]{*}{4} & Q     & Are there rivers or oceans that are not calm? \\
          & D     & [MASK]. \\
    \bottomrule
    \end{tabular}%
    \caption{Examples from the declaration generation on GQA test split. Q and D denote the question and generated declarative sentence by T5 model, respectively.}
  \label{tab:qtd_predictions}%
\end{table}%

Table \ref{tab:qtd_predictions_vqa} shows several declarative sentences generated from VQA v2.0 dataset. Since the questions in VQA v2.0 are raised manually, there exists various question types, which requires broader capabilities beyond the image understanding. As Table \ref{tab:qtd_predictions_vqa} shows, T5 model is able to generalize to VQA v2.0 dataset, and generate appropriate declarative sentences for most cases. For example, the question pattern in sample-1 (\ie \textit{what ... say?}) has not appeared in GQA dataset, but the T5 model still produces the proper declarative sentence. However, there also exist several question types that are difficult to convert, \eg \textit{why}, \textit{how}, etc. For example, the sample-2 asks the reason about \textit{the sailboats have their sails lowered}, but the [MASK] token in the declarative sentence is unable to prompt the answer. This limitation can explain the differences of the absolute accuracy improvement on GQA and VQA v2.0 datasets shown in Table \ref{tab:ablation_prompt}.

% Table generated by Excel2LaTeX from sheet 'Examples'
\begin{table}[t]
  \centering
    \begin{tabular}{|cll|}
    \toprule
    \textbf{Id} & \textbf{Type} & \textbf{Content} \\
    \midrule
    \multirow{2}[2]{*}{0} & Q     & What is the man doing? \\
          & D     & the man is [MASK]. \\
    \midrule
    \multirow{2}[2]{*}{1} & Q     & What does the sign say? \\
          & D     & the sign says the [MASK]. \\
    \midrule
    \multirow{2}[2]{*}{2} & Q     & Why do the sailboats have their sails lowered?  \\
          & D     & the sailboats have [MASK] lowered. \\
    \midrule
    \multirow{2}[2]{*}{3} & Q     & Could this roof be tiled? \\
          & D     & [MASK] \\
    \midrule
    \multirow{2}[2]{*}{4} & Q     & What pattern is the person's shirt? \\
          & D     & the shirt is [MASK]. \\
    \bottomrule
    \end{tabular}%
    \caption{Examples from the declaration generation on VQA v2.0 dataset. Q and D denote the question and generated declarative sentences by T5 model, respectively.}
  \label{tab:qtd_predictions_vqa}%
\end{table}%

\subsection{Visual Question Answering}

\begin{figure*}[t]
  \centering
  \includegraphics[width=1.\textwidth]{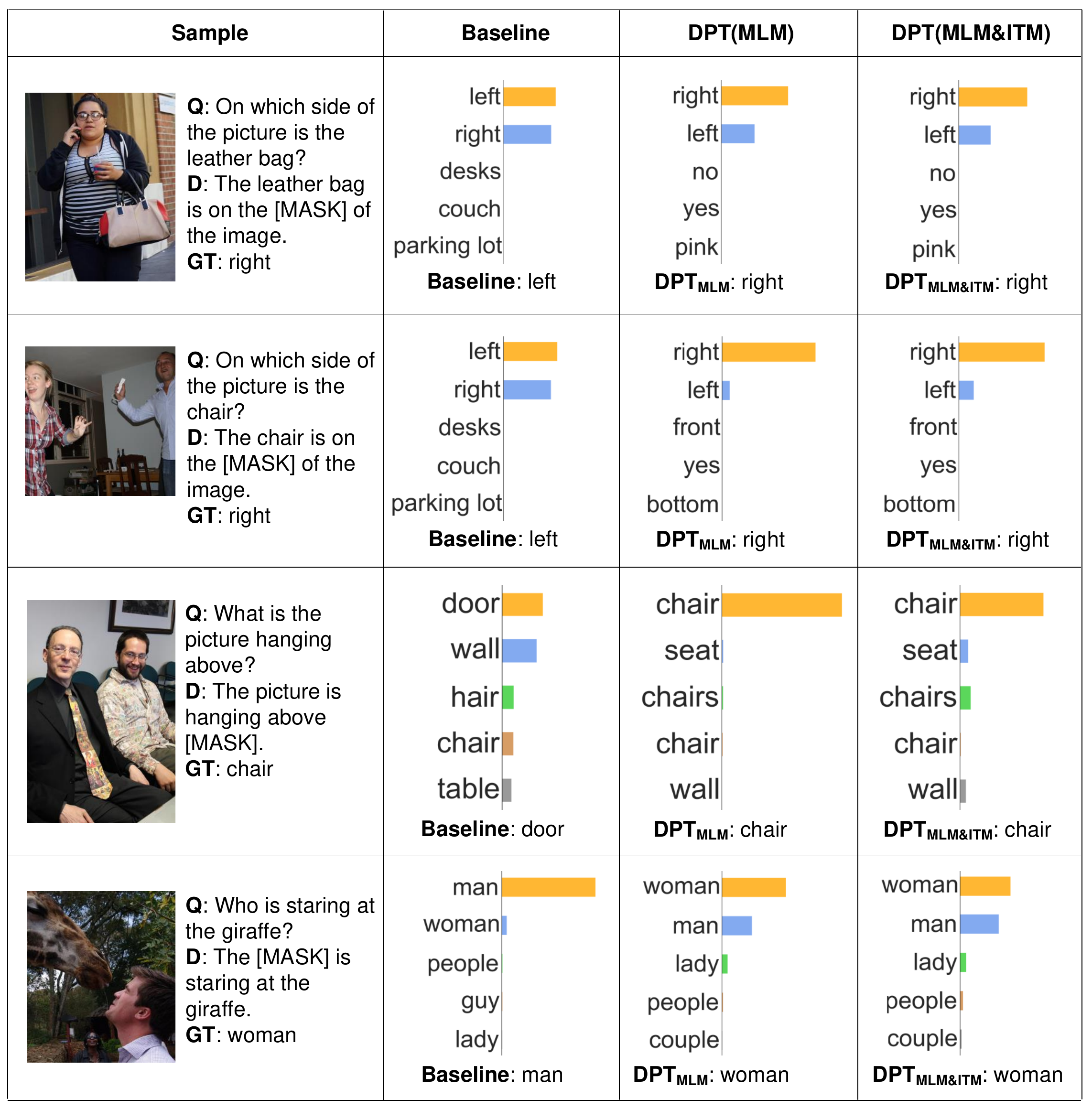} % Reduce the figure size so that it is slightly narrower than the column.
  \caption{Visualization of \textit{testdev} results on the GQA dataset. Top-5 answers predicted by Baseline, DPT(MLM) and DPT(MLM\&ITM) models are displayed in the right three columns.}
  \label{fig:cases_appendix_wrr}
\end{figure*}

In Figure \ref{fig:cases_appendix_wrr}, we visualize several prediction results from Baseline, DPT(MLM) and DPT(MLM\&ITM) models. The samples are taken from the cases where Baseline predicts wrong answers while our proposed DPT gets the correct ones. From the first two rows, it can be observed that Baseline fails on the direction-related questions, especially \textit{right/left} questions. The predicted scores of \textit{right} and \textit{left} from Baseline are almost equal, indicating that the model has difficulty in judging the directions. In contrary, our proposed DPT reserves a [MASK] token for answer cloze-filling. Benefit from the capability learned from the pre-training phase (\ie MLM), DPT is able to predict the correct answer with high confidence. The last two rows shows examples in which the [MASK] token indicates an object in the image. While the Baseline predicts wrong answers (\ie \textit{door} and \textit{man}), our proposed DPT can refer to the target objects in images, resulting in correct answers (\ie \textit{chair} and \textit{woman}).

\begin{figure*}[t]
  \centering
  \includegraphics[width=1.\textwidth]{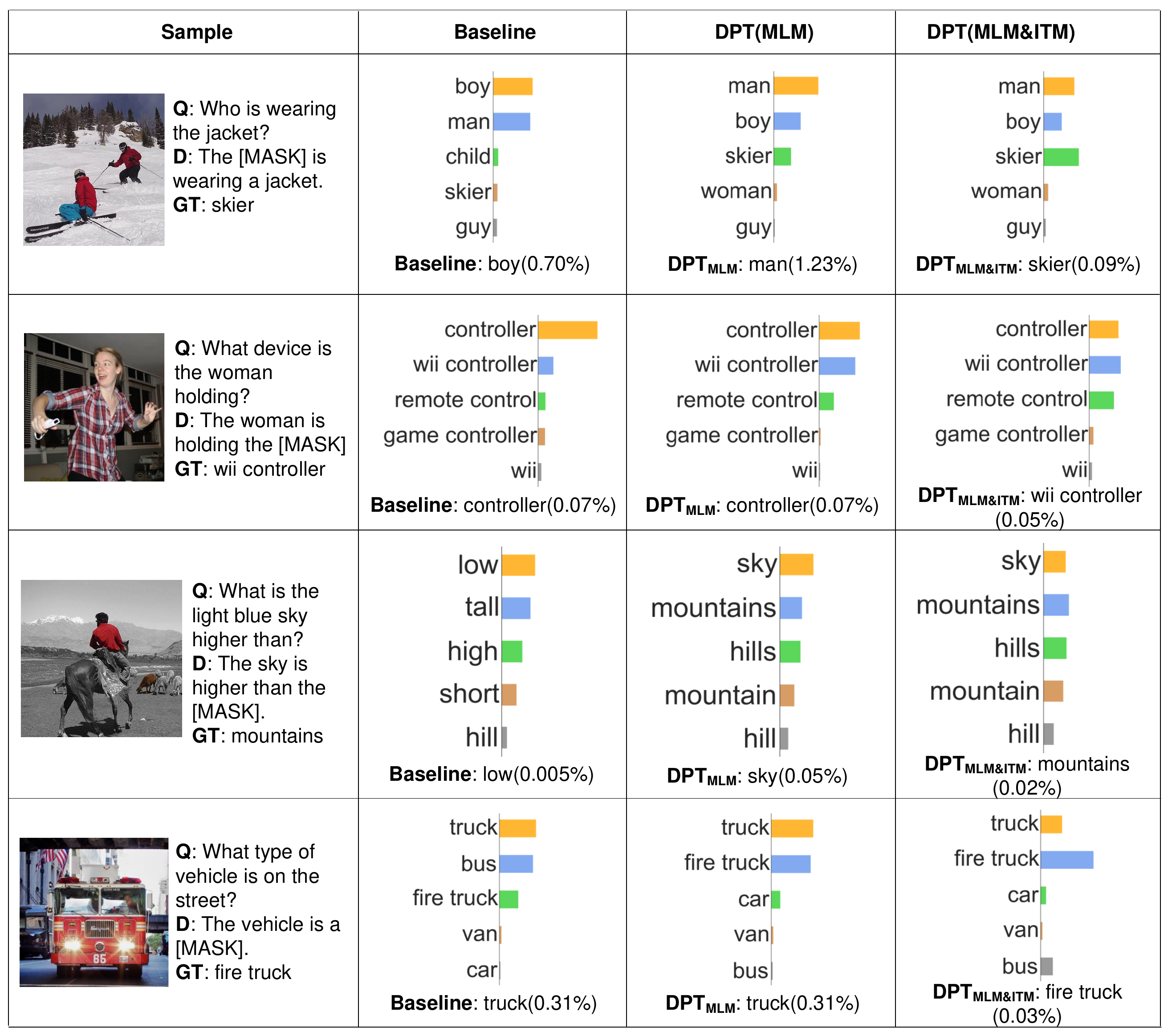} % Reduce the figure size so that it is slightly narrower than the column.
  \caption{Visualization of \textit{testdev} results on the GQA dataset. Top-5 answers predicted by Baseline, DPT(MLM) and DPT(MLM\&ITM) models are displayed in the right three columns. The percentage in parentheses denotes the proportion of answers in the training set.}
  \label{fig:cases_appendix_wwr}
\end{figure*}

Figure \ref{fig:cases_appendix_wwr} visualizes the samples where Baseline and DPT(MLM) predict wrong answers while DPT(MLM\&ITM) gets the correct ones. As shown in the examples, the Baseline and DPT(MLM) tend to predict general answers that appear more offen in the training set, \eg \textit{boy}, \textit{man}, \textit{truck} \etc Such bias is caused by the classification layer in the model. On the contrary, our proposed DPT with ITM task chooses the answer via the matching scores, instead of the answer probabilities. Therefore, our proposed method is able to mitigate the answer prediction bias to a extent, and predicts more granular answers due to the consideration of answer semantics, \eg \textit{controller} vs. \textit{wii controller}, \textit{truck} vs. \textit{fire truck}.

\subsection{Attention Visualization}

\begin{figure*}[t]
  \centering
  \includegraphics[width=1.\textwidth]{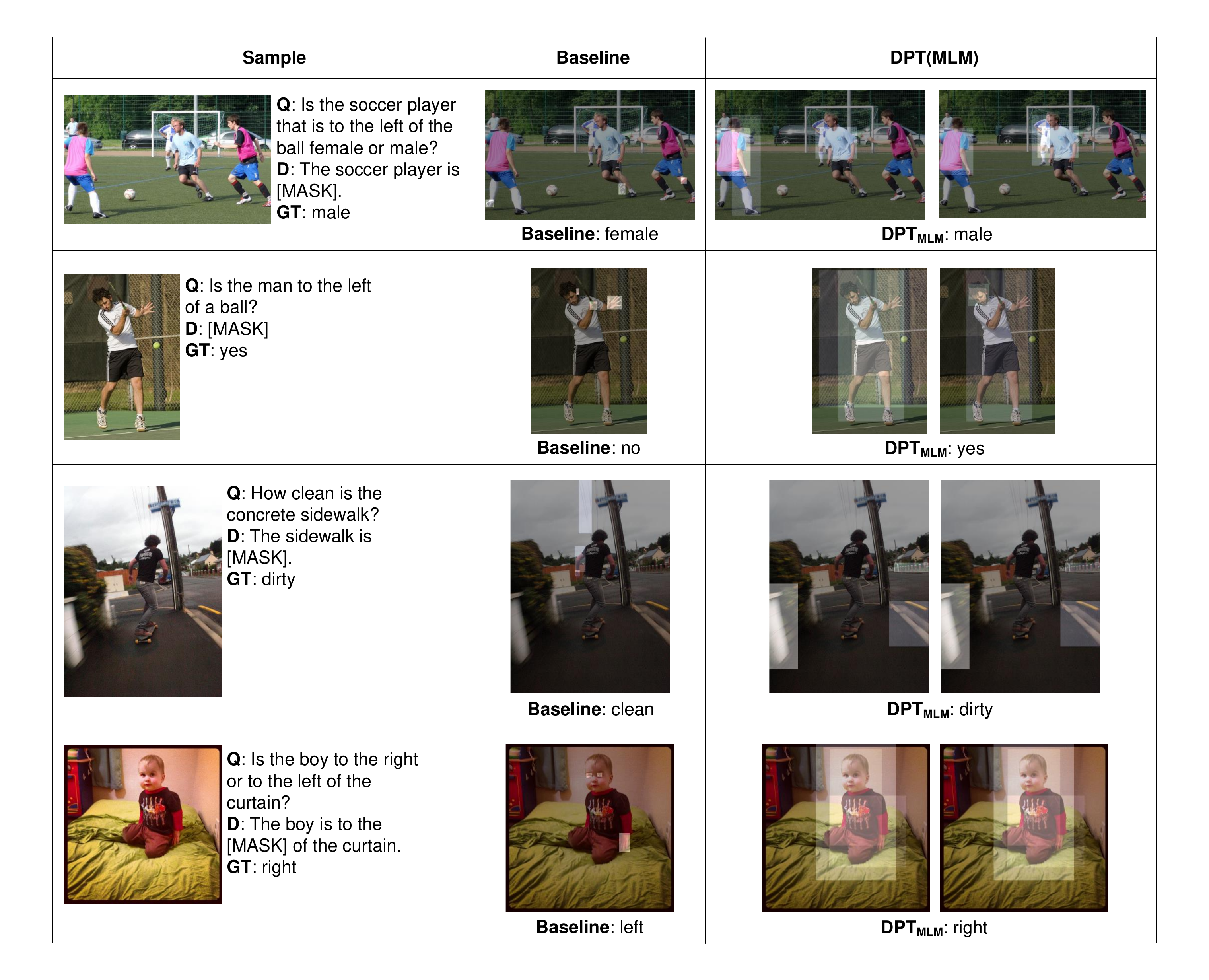} % Reduce the figure size so that it is slightly narrower than the column.
  \caption{Visualization of attention maps corresponding to [CLS] and [MASK] tokens. All attention maps are obtained from the last Transformer layer. Top-3 objects with the maximum attention scores are visualized in the image. As for Baseline, the attention map from [CLS] is exploited for visualization. As for DPT(MLM), the two images show the attention maps of [CLS] (left) and [MASK] (right) tokens, respectively.}
  \label{fig:attention}
\end{figure*}

In Figure \ref{fig:attention}, we visualize the attention maps corresponding to [CLS] and [MASK] tokens from the last Transformer layer. From these examples, it can be observed that in some cases, the [CLS] token learned from Baseline model fails to focus on the question-relevant regions, resulting in wrong answers. On the contrary, the learned [CLS] and [MASK] tokens show reasonable attentions on the relevant objects, thus producing the right answers.

\end{document}